%% file: main.tex
\documentclass[letterpaper, 10 pt, conference]{ieeeconf}  

\IEEEoverridecommandlockouts                              

\usepackage[T1]{fontenc}

\usepackage{amsfonts}	
\usepackage{amsmath}	
\usepackage{amssymb}    
\usepackage{siunitx}
\usepackage{xfrac}    
\usepackage{pifont}   

\usepackage{booktabs}
\usepackage{makecell}  
\usepackage[flushleft]{threeparttable}  
\usepackage{multirow}
\usepackage{rotating}

\usepackage{xspace}    
\usepackage[dvipsnames]{xcolor}    
\usepackage{colortbl}
\let\labelindent\relax
\usepackage[inline]{enumitem} 
\usepackage{graphicx} 
\usepackage{microtype}
\usepackage{cite}
\usepackage{flushend}

\usepackage{algorithm2e}

\makeatletter
\let\NAT@parse\undefined
\makeatother

\usepackage{url}

\usepackage[pdfencoding=auto, colorlinks=true]{hyperref} 
\usepackage[hang,flushmargin]{footmisc}

\usepackage[capitalize]{cleveref}
\crefname{section}{Sec.}{Secs.}
\Crefname{section}{Section}{Sections}
\Crefname{table}{Table}{Tables}
\crefname{table}{Tab.}{Tabs.}

\newcommand{\grayrule}{\arrayrulecolor{black!30}\midrule\arrayrulecolor{black}}
\definecolor{Gray}{gray}{0.9}

\newcommand{\method}{\mbox{L\textsuperscript{3}PS}\xspace}

\DeclareMathOperator*{\argmax}{arg\,max}

\definecolor{barrier}{RGB}{255, 120, 50} 
\definecolor{bicycle}{RGB}{100, 230, 245} 
\definecolor{bus}{RGB}{100, 80, 250} 
\definecolor{car}{RGB}{100, 150, 245} 
\definecolor{construction}{RGB}{100, 149, 237} 
\definecolor{motorcycle}{RGB}{30, 60, 150} 
\definecolor{pedestrian}{RGB}{255, 30, 30} 
\definecolor{traffic}{RGB}{50, 255, 255} 
\definecolor{trailer}{RGB}{0, 0, 255} 
\definecolor{truck}{RGB}{80, 30, 180} 
\definecolor{driveable}{RGB}{255, 0, 255} 
\definecolor{otherflat}{RGB}{175, 0, 75} 
\definecolor{sidewalk}{RGB}{75, 0, 75} 
\definecolor{terrain}{RGB}{150, 240, 80} 
\definecolor{manmade}{RGB}{255, 200, 0} 
\definecolor{vegetation}{RGB}{0, 175, 0} 

\begin{document}

\title{\LARGE \bf
Label-Efficient LiDAR Panoptic Segmentation
}

\author{
Ahmet Selim Çanakçı$^{1*}$, 
Niclas Vödisch$^{1*}$, 
Kürsat Petek$^{1}$, 
Wolfram Burgard$^{2}$, 
and Abhinav Valada$^{1}$
\thanks{$^{*}$ Equal contribution.}%
\thanks{$^{1}$ Department of Computer Science, University of Freiburg, Germany.}%
\thanks{$^{2}$ Department of Eng., University of Technology Nuremberg, Germany.}%
\thanks{This work was funded by the German Research Foundation (DFG) Emmy Noether Program grant number 468878300.}%
}

\maketitle
\thispagestyle{empty}
\pagestyle{empty}


\begin{abstract}
    \input{sections/0_abstract}
\end{abstract}


\input{sections/1_introduction}

\input{sections/2_related_work}

\input{sections/3_method}
\input{sections/4_experiments}
\input{sections/5_conclusion}



\footnotesize
\bibliographystyle{IEEEtran}
\bibliography{references.bib}


\end{document}

%% file: sections/0_abstract.tex
A main bottleneck of learning-based robotic scene understanding methods is the heavy reliance on extensive annotated training data, which often limits their generalization ability. In LiDAR panoptic segmentation, this challenge becomes even more pronounced due to the need to simultaneously address both semantic and instance segmentation from complex, high-dimensional point cloud data. In this work, we address the challenge of LiDAR panoptic segmentation with very few labeled samples by leveraging recent advances in label-efficient vision panoptic segmentation. To this end, we propose a novel method, Limited-Label LiDAR Panoptic Segmentation (\method), which requires only a minimal amount of labeled data. Our approach first utilizes a label-efficient 2D network to generate panoptic pseudo-labels from a small set of annotated images, which are subsequently projected onto point clouds. We then introduce a novel 3D refinement module that capitalizes on the geometric properties of point clouds. By incorporating clustering techniques, sequential scan accumulation, and ground point separation, this module significantly enhances the accuracy of the pseudo-labels, improving segmentation quality by up to $+10.6$~PQ and $+7.9$~mIoU. We demonstrate that these refined pseudo-labels can be used to effectively train off-the-shelf LiDAR segmentation networks. Through extensive experiments, we show that \method not only outperforms existing methods but also substantially reduces the annotation burden. We release the code of our work at \url{https://l3ps.cs.uni-freiburg.de}.

%% file: sections/1_introduction.tex
\section{Introduction}

LiDAR panoptic segmentation~\cite{milioto2020lps} enables robots to perceive and interpret their 3D environment by segmenting both instance and semantic information. However, most existing approaches~\cite{sirohi2021efficientlps,zhou2021polarnet,xiao2024position} rely on large, often human-annotated, training datasets. Obtaining such data is both time-consuming and labor-intensive due to the inherent complexity of 3D data and the requirement for point-wise annotation precision. While learning-based tasks in other domains~\cite{oquab2023dinov2, touvron2023llama, oneill2024openx} have greatly benefited from the vast availability of annotated datasets, the 3D domain still suffers from a lack of comparable amount of training data. Moreover, generalization in the 3D space is considerably more challenging than in the 2D visual domain due to more multifaceted domain gaps, including semantic and structural variations. Consequently, developing a foundation model for 3D representations remains an open quest~\cite{wu2024ppt}, highlighting the importance of label efficiency in robotic 3D perception.

\begin{figure}[t]
    \centering
    \includegraphics[width=\linewidth]{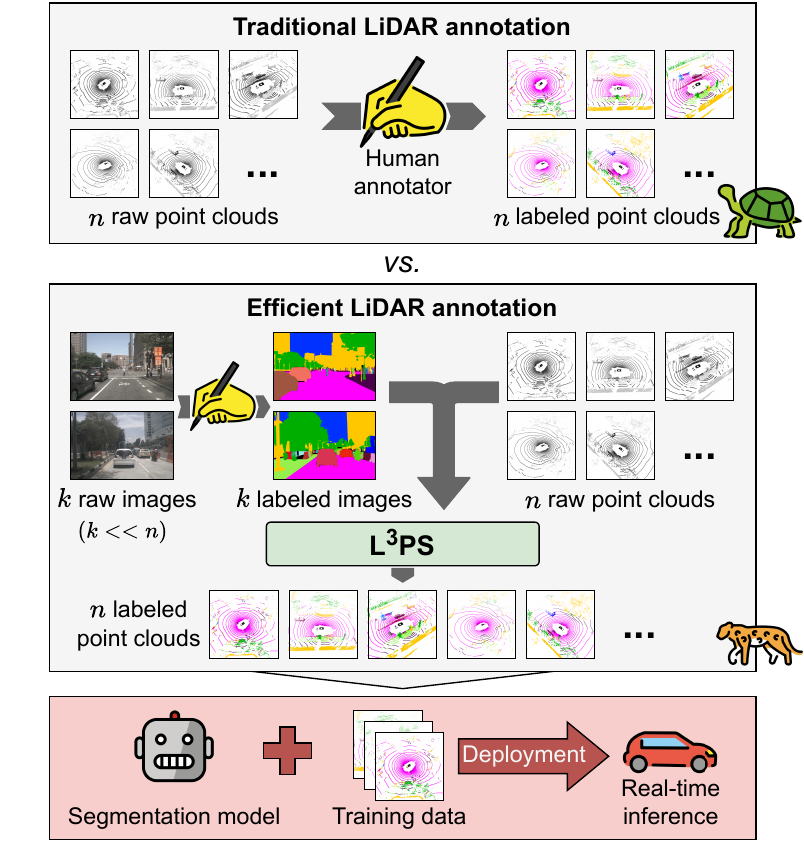}
    \vspace*{-.6cm}
    \caption{
    Most learning-based methods for LiDAR panoptic segmentation require a large amount of annotated point clouds for training. Generating this data is both expensive and time-consuming. We present a more efficient alternative that reduces the annotation effort from $n$~point clouds to $k$~images, which a human annotator can process within a single day. Our proposed \method approach combines recent advances from label-efficient visual panoptic segmentation with spatial information from the LiDAR scans to generate high-quality 3D panoptic pseudo-labels. We demonstrate the utility of these labels to train off-the-shelf LiDAR segmentation networks.
    }
    \label{fig:teaser}
    \vspace*{-.6cm}
\end{figure}

To address the challenges associated with LiDAR panoptic segmentation, some studies~\cite{bevsic2022unsupervised} have explored unsupervised domain adaptation, aiming to transfer knowledge from a labeled source domain to an unlabeled target domain. However, unlike in the image domain, where adaptation techniques~\cite{voedisch23codeps, saha2023edaps} have achieved significant success, their application in the 3D space is considerably more complex. Consequently, most existing approaches rely on fully annotated point cloud datasets or focus on specific sub-tasks within panoptic segmentation. In contrast, recent vision-based methods have demonstrated the potential of label-efficient techniques, such as self-supervised learning~\cite{oquab2023dinov2, gosala2024letsmap, gosala2023skyeye} and pretraining on large-scale datasets~\cite{kappeler2024spino, voedisch2025pastel}. These approaches substantially reduce the annotation burden while maintaining competitive performance relative to their fully supervised counterparts.

In this work, we propose a novel method, called \textbf{L}imited-\textbf{L}abel \textbf{L}iDAR \textbf{P}anoptic \textbf{S}egmentation (\method) that addresses label efficiency in terms of annotation costs. We argue that generating panoptic annotations for 2D images is substantially less expensive than for 3D point clouds. Therefore, our \method approach leverages recent 2D label reduction techniques~\cite{voedisch2025pastel} to generate large-scale 2D pseudo-labels from a minimal set of images. This is followed by a multi-step enhancement process that produces panoptic point cloud annotations. We demonstrate that the quality of these point cloud annotations is sufficient to effectively train off-the-shelf LiDAR segmentation networks.

The main contribution of this paper is the proposed \method approach that addresses the challenge of label-efficient LiDAR panoptic segmentation. We illustrate the key idea in \cref{fig:teaser}. In this paper, we make three claims:
\begin{enumerate*}[label={(\arabic*)}]
    \item Our method substantially reduces the reliance on expensive panoptic point cloud annotations by leveraging visual foundation models for image-informed point pseudo-labels.
    \item Our proposed post-processing module enhances the pseudo-label quality by exploiting the geometric properties of point clouds.
    \item Overall, \method achieves competitive performance compared to existing baselines while requiring less annotation effort.
\end{enumerate*}
We validate these claims through extensive experiments on the panoptic nuScenes dataset~\cite{fong2022panoptic} and comprehensively analyze architectural design choices, demonstrating their impact on panoptic segmentation quality. Our results show our method's effectiveness in advancing LiDAR panoptic segmentation with minimal annotation effort. To facilitate reproducibility and future research, we open-source \method on \mbox{\url{https://l3ps.cs.uni-freiburg.de}}.

%% file: sections/2_related_work.tex
\section{Related Work}
In this section, we present an overview of LiDAR panoptic segmentation and label-efficient segmentation of point clouds.


{\parskip=2pt\noindent\textit{LiDAR Panoptic Segmentation:}
LiDAR panoptic segmentation~\cite{milioto2020lps} aims to classify each point in a LiDAR point cloud into its corresponding semantic category (e.g., \textit{car}, \textit{pedestrian}, \textit{road}) while simultaneously grouping points that belong to the same object instance (e.g., different cars). This is particularly challenging due to occlusions, varying object sizes, and the inherent sparsity of LiDAR data. To address these challenges, initial approaches~\cite{sirohi2021efficientlps, milioto2020lps} process LiDAR data using range images and employ a network design inspired by image-based architectures.
An alternative is to process LiDAR data directly on the raw point cloud. Methods such as PointPillars~\cite{lang2017pointpillars} generate pseudo-images from vertically organized point clouds, whereas VoxelNet~\cite{zhou2018voxelnet} applies 3D convolutions to voxelized point clouds. Cylinder3D\cite{zhu2021cylindrical}  extends this idea by voxelizing the point cloud, transforming it into a cylindrical coordinate system, partitioning the space into non-uniform grids, and applying asymmetrical convolutions.} 

In contrast, Panoster~\cite{gasperini2021panoster} operates directly on raw point clouds, introducing the first end-to-end LiDAR panoptic segmentation network by predicting instance IDs for each point. Panoptic-PolarNet~\cite{zhou2021polarnet} improves on this method by introducing a novel encoder that transforms the input point cloud into a polar bird's-eye-view frame while remaining fully end-to-end trainable. Furthermore, Panoptic-PHNet~\cite{li2022panoptic} merges bird's-eye-view transformation with the common voxel encoding to improve the precision. One of the recent approaches, MostLPS~\cite{agarwalla2023lidar} follows a segmentation-centric fashion with a detection-centric design, leveraging trajectory-level supervision to infer object size and associating LiDAR points for fine-grained instance segmentation. On the other hand, P3Former~\cite{xiao2024position} processes a voxelized point cloud using Cylinder3D~\cite{zhu2021cylindrical} for feature encoding and incorporates learnable queries that are processed along with backbone features in a multi-layer segmentation head. 
A drawback of these traditional methods is their reliance on a large set of annotated data. Hence, in this work, we demonstrate successful training using the label-efficient pseudo-labels generated by our approach.


\begin{figure*}[t]
    \centering
    \includegraphics[width=\linewidth]{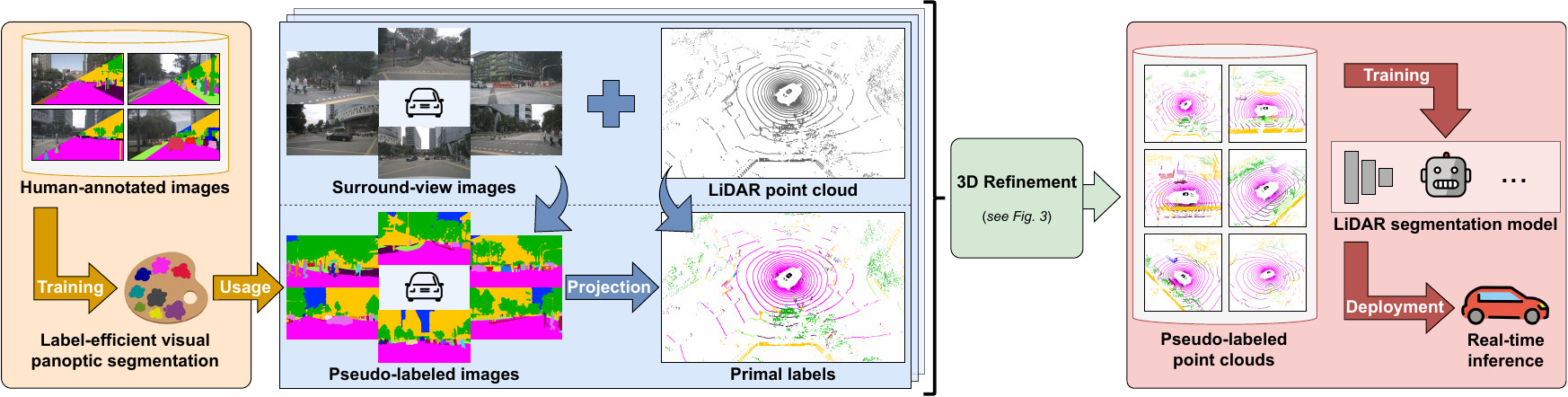}
    \vspace*{-.7cm}
    \caption{Overview of our proposed \method approach for label-efficient LiDAR panoptic segmentation. First, we train a label-efficient image-based panoptic segmentation model using a minimal number of human-annotated images. Second, we utilize the trained model to generate pseudo-labels for surround-view image data, which are then projected onto the corresponding LiDAR point clouds. Third, we employ our proposed 3D refinement module. Finally, we use the refined point cloud pseudo-labels to train an off-the-shelf LiDAR segmentation model, which can then be deployed for real-time inference.}
    \label{fig:overview}
    \vspace*{-.4cm}
\end{figure*}


{\parskip=2pt
\noindent\textit{Label-Efficient Point Cloud Segmentation:}
Manual labeling for LiDAR segmentation is time-consuming and expensive~\cite{wang2019latte}, limiting the practicality of traditional approaches. To overcome this challenge, label-efficient methods aim to match the performance of fully supervised methods while significantly reducing the need for annotated data. Existing methods can generally be categorized into self-supervised approaches~\cite{xie2020pointcontrast, hou2021exploring, zhang2023growsp, liu2024u3ds3}, which do not require human annotations during training, and weakly supervised approaches~\cite{liu2023cpcm, xu2023hpal}. These approaches rely on a reduced amount of labeled data, either in quantity (e.g., 10 labels vs. 1000), as proposed in our work, or the form of supervision (e.g., single points versus entire point clouds). Initial self-supervised approaches~\cite{xie2020pointcontrast, hou2021exploring} employ contrastive learning-based pretraining to extract point discriminators for clustering and use only a small fraction of the training data for segmentation. Other methods leverage clusters of spatially proximate, semantically similar points (``superpoints''). In GrowSP~\cite{zhang2023growsp}, superpoints are formed by progressively expanding per-point neighborhoods to promote intra-cluster feature similarity and inter-cluster distinctiveness, whereas U3DS3~\cite{liu2024u3ds3} generates superpoints based on geometric structure, such as surface normals. Although these techniques do not require any training annotations, an alternative is to reduce the labeling effort by using a single point annotation per instance~\cite{liu2023cpcm} or by predicting point uncertainties~\cite{xu2023hpal} to select points for manual labeling.} 

Recently, image-based foundation models such as \mbox{DINOv2}~\cite{oquab2023dinov2}, SAM~\cite{kirillov2023sam}, and CLIP~\cite{radford2021clip} have been widely adopted to project their features or segmentation masks onto point clouds. The use of these vision foundation models in label-efficient point cloud segmentation can be categorized into knowledge distillation from 2D to 3D~\cite{sautier2022image, chen2023pointdc, mahmod2023self, puy24scalr, liu2024seal} and direct application via projection and label generation~\cite{kweon2024real, ovsep2025sal}. These models primarily extract image features using a vision foundation model, e.g., SL, with SLidR~\cite{sautier2022image} projecting image superpixels into 3D and learning through a superpixel-driven contrastive distillation loss. Alternatively, PointDC~\cite{chen2023pointdc} distills features from multiple scene views, while ST-SLidR~\cite{mahmod2023self} integrates both superpixel and superpoint features, employing a semantically tolerant contrastive loss between matching superpoint-pixel pairs. Moreover, ScaLR~\cite{puy24scalr} pretrains a single backbone on a mixture of datasets prior to distillation, introducing a significant performance boost. Lastly, SEAL~\cite{liu2024seal} employs a contrastive approach across modalities and time by generating superpixels from multiple diverse visual foundation models, which are projected to 3D to form superpoints. These are encoded separately in 2D and 3D encoders, while spatially contrasting camera and LiDAR data and temporally contrasting LiDAR point clouds. 

In contrast, unlike other methods that distill knowledge from 2D foundation models to 3D backbones, REAL~\cite{kweon2024real} iteratively refines segmentation masks by projecting weakly supervised 3D points into 2D, generating masks with SAM~\cite{kirillov2023sam}, and back-projecting them into 3D. This is combined with CLIP~\cite{radford2021clip} in SAL~\cite{ovsep2025sal} to further support zero-shot segmentation of point clouds. The recent LeAP~\cite{gebraad2025leap} approach uses vision foundation models to generate zero-shot 2D semantic labels. These labels are then transferred via a voxel-fusion module and a confidence-based 3D encoder, followed by neighborhood smoothing. Despite these advancements, existing label-efficient methods exhibit one or more of the following limitations: (1)~they fail to achieve competitive performance, (2)~they rely on vision foundation models that require human supervision, and (3)~they depend heavily on point cloud annotations. In this work, we address these challenges by substituting resource-intensive point cloud annotations with fewer, more cost-effective image annotations.

%% file: sections/3_method.tex
\section{Technical Approach}

In this section, we first define the problem addressed in this work and provide an overview of our approach. We then detail the individual steps and conclude by describing an application scenario for our method.


\begin{figure*}[t]
    \centering
    \includegraphics[width=\linewidth]{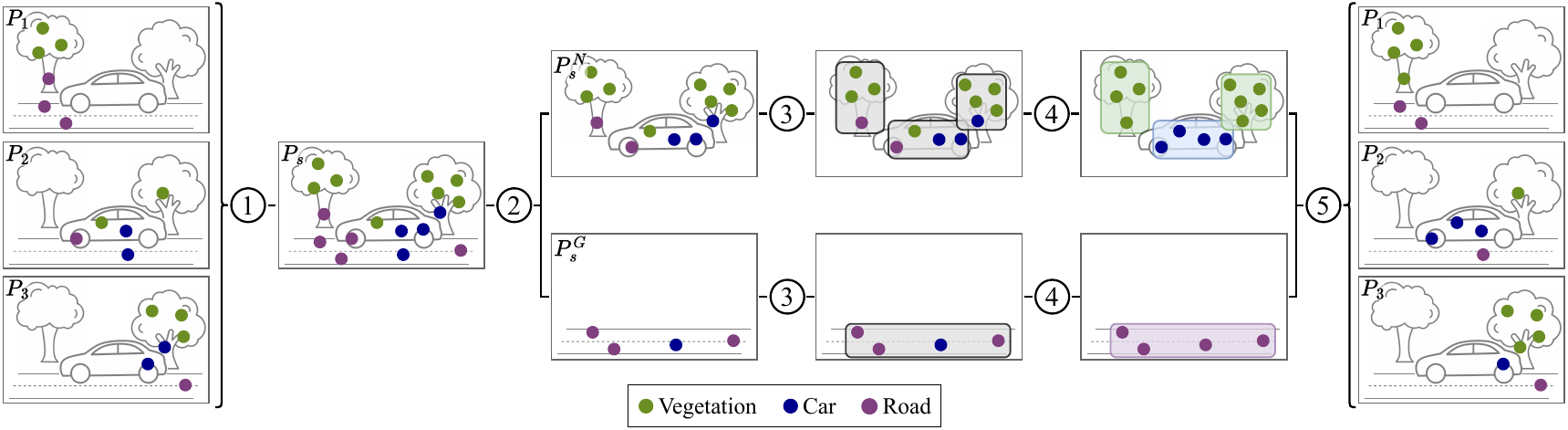}
    \vspace*{-.6cm}
    \caption{For 3D refinement, we employ the following steps: (1)~We accumulate the individual scans $\{P_1, P_2, P_3\}$ in a single scene $P_s$. (2)~Afterwards, we separate the scene into a ground partition $P_s^G$ and a non-ground partition $P_s^N$. (3)~Next, we cluster the points in both partitions based on their 3D position. Ideally, this should separate instances. (4)~In each cluster, we perform majority voting on the semantic classes. (5)~Finally, we separate the scene into the original LiDAR scans and propagate the class corrections to the instance labels. For simplicity, we omit the instance labels in this visualization.
    }
    \label{fig:3d-refinement}
    \vspace*{-.3cm}
\end{figure*}


\subsection{Problem Definition}
In this work, we consider the following scenario:
Given is a set of $n$ point clouds $\mathcal{P} = \{P_1, \dots, P_n\}$, where each point cloud is composed of 3D points $p_i \in \mathbb{R}^3$, and a set of corresponding surround-view images $\mathcal{I} = \{I_1, \dots, I_n \}$. Assume that for a small subset of the images $\mathcal{I}_A \subset \mathcal{I}$, panoptic annotations are available. The goal is to bootstrap these annotations to generate panoptic point cloud pseudo-labels for the entire set $\mathcal{P}$. The core problem is that due to $|\mathcal{I}_A| \ll n$, transferring the annotations from the image domain to the point clouds is not trivial. Furthermore, classical methods that require a large amount of training data are not applicable.


\subsection{Overview}
The core concept of our approach is to leverage vision foundation models that produce high-quality labels from a limited number of labeled samples, which are projected onto point clouds and refined through a dedicated post-processing module. An overview of the pipeline is shown in \cref{fig:overview}.
First, we train a label-efficient visual panoptic segmentation model using only a minimal number of human-annotated images, thereby reducing the annotation effort compared to point cloud labeling. Second, we use this model to generate pseudo-labels for surround-view image data, which are subsequently projected onto the corresponding LiDAR point clouds. Third, we employ our proposed 3D refinement module, which accumulates individual LiDAR scans into a consistent scene representation and performs various post-processing steps by exploiting spatial information. Finally, we use the refined point cloud pseudo-labels to train off-the-shelf LiDAR segmentation models, enabling real-time inference.


\subsection{Primal Label Generation}
\label{ssec:primal-label-generation}
In the initial step, we use the annotations of $\mathcal{I}_A$ to obtain pseudo-labels for the unlabeled images in $\mathcal{I} \setminus \mathcal{I}_A$.
For this task, we employ the recent label-efficient image panoptic segmentation method PASTEL~\cite{voedisch2025pastel} that comprises two lightweight heads for semantic segmentation and object delineation on top of a frozen DINOv2~\cite{oquab2023dinov2} backbone.
We predict pseudo-labels for the remaining images after training PASTEL using images $\mathcal{I}_A$ and ground truth annotations.
Finally, we project each point cloud $P_j$ onto the corresponding surround-view image $I_j$ and annotate every point with the 2D panoptic pseudo-label of the target pixel. Formally, we apply the projection function $\Pi_j(\cdot)$ to a point $p_i \in P_j$ to obtain the 2D image coordinates $p_i'$: \looseness=-1
\begin{equation}
    p_i' = \Pi_j (p_i)
\end{equation}
Assuming that $p_i'$ is on the image plane of $I_j$, the semantic and instance pseudo-labels $\hat{y}_i \in \mathcal{Y}$ and $\hat{z}_i \in \mathbb{Z}$ of $p_i$ are defined as:
\begin{align}
    \hat{y}_i &= \text{class}(p_i') \, , \\
    \hat{z}_i &= \text{instance}(p_i') \, ,
\end{align}
where $\text{class}(\cdot)$ denotes the 2D semantic annotation of a pixel in $I_j$ and $\text{instance}(\cdot)$ is the corresponding instance ID. Points that are not in the field-of-view of any surround-view camera are assigned a special \textit{void} label at this stage.
Due to the multi-step structure of our approach, we refer to the initial projection-based pseudo-labels $(\hat{y}, \hat{z})$ as primal labels.


\subsection{3D Refinement}
\label{ssec:3d_refinement}
In the subsequent steps, we leverage the 3D information of a point cloud to further enhance the primal labels. Throughout this section, we refer to the steps visualized in \cref{fig:3d-refinement}.

{\parskip=2pt \noindent\textit{Accumulation and Ground Segmentation:}
First, we compute relative 3D poses for temporally contiguous LiDAR scans based on KISS-ICP~\cite{vizzo2023ral}. Then, we use these poses to accumulate the contiguous scans in larger point clouds $\mathcal{P}_S = \{ P_{s_1}, \dots, P_{s_n} \}$, where we refer to an accumulated point cloud $P_s \in \mathcal{P}_S$ as an individual scene (\textit{step 1}). Note that this nomenclature follows the common sequence-oriented setup in autonomous driving datasets~\cite{caesar2020nuscenes,behley2019semantickitti}.
Second, we generate ground segmentation masks using Patchwork++~\cite{lee2022patchwork++} to separate a scene $P_s$ into a ground partition $P_s^G$ and a non-ground partition $P_s^N$ (\textit{step 2}). These partitions are processed individually in the subsequent steps.
}

{\parskip=2pt \noindent\textit{Clustering and Correction:}
For each scene, we extract point clusters $\mathcal{C}_s = \{C_1, \dots, C_{m}\}$ separately for both partitions $P_s^G$ and $P_s^N$ using HDBSCAN~\cite{campello2013density} (\textit{step 3}). Furthermore, all points that were not assigned to any cluster by HDBSCAN are collected in an additional cluster $C_\textit{noise}$. We reassign all points in $C_\textit{noise}$ to their respective nearest cluster in $\mathcal{C}_s$ using a \mbox{$k$-nearest} neighbors classifier. Intuitively, the geometry-based clusters should approximate semantic objects, i.e., all points within a cluster should receive the same pseudo-label (\textit{step 4}).
We further define a set of rare semantic classes $\mathcal{Y}^R \subset \mathcal{Y}$ that are underrepresented in the primal labels, e.g., \textit{construction vehicle}. For each cluster, we compute the frequency of each semantic class. If the most frequent class is \textit{void} and the frequency is greater than a threshold $\tau_{\mathit{void}}$, we assign all points in the cluster to \textit{void}. If the frequency of one or multiple rare classes is above a threshold $\tau$, we use the most frequent rare class as the pseudo-label of all points within the cluster. Otherwise, we use the cluster's most frequent label:
\begingroup
\setlength{\thinmuskip}{2.15mu} 
\setlength{\thickmuskip}{1.0mu}
\begin{align}
    y^*_i = 
    \begin{cases}
        y, \hspace*{-5pt} & \text{if } \exists y \in \mathcal{Y}^R : \text{freq}(y) > \tau\\
        \argmax\limits_y \{\text{freq}(y) \, | \, y \in \mathcal{Y}\}, \hspace*{-5pt} & \text{otherwise}
    \end{cases}
\end{align}
\endgroup
Here, $\text{freq}(y)$ is the in-cluster frequency of a semantic label~$y$ and the updated pseudo-label of a point $p_i$ is denoted by $y^*_i$.
}

{\parskip=2pt \noindent\textit{Instance Correction:}
Until now, we have only updated the semantic class of the panoptic pseudo-labels without considering the instance information. In the final step, we transfer the previous semantic changes to the instance labels. Unlike the previous steps, in which we operated on the accumulated scenes $\mathcal{P}_S$, we now correct the instances on the original point clouds $\mathcal{P}$ (\textit{step 5}). For every point $p_i \in P_j$ of a given point cloud $P_j \in \mathcal{P}$, we first assign an instance label~$z^*_i$ via: \looseness=-1
\begin{align}
    z^*_i =
    \begin{cases}
        \hat{z}_i, & \text{if } \hat{y}_i = y^*_i \\
        \textit{unknown}, & \text{otherwise}
    \end{cases}
    \label{eqn:instance-correction}
\end{align}
Afterward, we resolve the \textit{unknown} instance IDs by adopting the instance ID of the point's nearest neighbor that was assigned a valid ID in \cref{eqn:instance-correction}. Note that as defined by the panoptic segmentation task, points belonging to the semantic \textit{stuff} class are not assigned an instance ID.
}


\subsection{Segmentation Network Training}

After generating pseudo-labels for unlabeled point cloud data, we train off-the-shelf LiDAR semantic and panoptic segmentation models with them as illustrated in~\cref{fig:overview}. Specifically, we train ScaLR~\cite{puy24scalr} for semantic segmentation and P3Former~\cite{xiao2024position} for panoptic segmentation. These methods also allow for real-time point cloud segmentation.

{\parskip=2pt \noindent\textit{Semantic Segmentation:}
ScaLR~\cite{puy24scalr} distills task-agnostic features from vision foundation models to the LiDAR domain. For semantic segmentation, the network requires fine-tuning with semantic point cloud annotations. In this work, we substitute the normally used ground truth labels with our generated pseudo-labels and fine-tune the ScaLR model with the following loss:
\begin{align}
    \mathcal{L} = \mathcal{L}_{CE}(\mathbf{p}, \mathbf{y}) + \lambda \cdot \mathcal{L}_{\text{Lovász}} \left( \text{softmax}(\mathbf{p}, \mathbf{y}) \right) \, ,
\end{align}
with $\mathcal{L}_{CE}(\mathbf{p}, \mathbf{y})$ denoting the cross entropy loss:
\begin{align}
    \mathcal{L}_{CE}(\mathbf{p}, \mathbf{y}) = - \sum_{i \in \mathcal{I}} \log p_{i, y_i} \, ,
\end{align}
where $ p_{i, y_i} $ is the predicted probability for the correct class $ y_i $ of pixel $ i $, and $ \mathcal{I} $ is the set of all pixels. $\mathcal{L}_{\text{Lovász}} $ is the Lovász softmax loss applied only to valid points ($y \neq \textit{void/ignore}$) with weighting factor $\lambda$.
}

{\parskip=2pt \noindent\textit{Panoptic Segmentation:}
In contrast, P3Former~\cite{xiao2024position} adheres to the conventional framework of supervised learning. We train the model from scratch using our generated panoptic pseudo-labels, with the following loss function:
\begin{align}
    \mathcal{L} = \lambda_c \mathcal{L}_c + \lambda_{fs} \mathcal{L}_{fs} + \lambda_{ps} \mathcal{L}_{ps}
\end{align}
The total loss $\mathcal{L}$ is composed of a classification loss~$\mathcal{L}_c$, a feature-seg loss~$\mathcal{L}_{fs}$, and a position-seg loss~$\mathcal{L}_{ps}$. Each component is multiplied by a corresponding weighting factor~$\lambda$. For further details, we refer to P3Former~\cite{xiao2024position}.
}

%% file: sections/4_experiments.tex
\section{Experimental Evaluation}

In this section, we first detail the experimental setup and evaluation protocol. Subsequently, we present extensive experiments supporting our key claims, namely (1)~\method reduces the reliance on expensive point cloud annotations, (2)~its post-processing module effectively improves the quality of the generated pseudo-labels by leveraging structural information, and (3)~\method achieves competitive performance compared to existing baselines while requiring less annotation effort.\looseness=-1


\subsection{Dataset and Evaluation Metrics}

To evaluate our method, we require a dataset that provides synchronized LiDAR scans and surround-view images. Although this setup is common in mobile robotics, ground truth panoptic annotations are rarely available for both modalities in the same dataset. While LiDAR annotations are required for evaluation, a small number of image annotations are required to train the label-efficient image segmentation network. The most widely used benchmarks~\cite{li2022panoptic,agarwalla2023lidar,xiao2024position,mahmod2023self,liu2024seal,puy24scalr} for LiDAR panoptic segmentation are Panoptic nuScenes~\cite{fong2022panoptic} and SemanticKITTI~\cite{behley2019semantickitti}. Neither of these datasets provides annotated surround-view images. However, the nuImages dataset~\cite{caesar2020nuscenes} was recorded in the same domain as nuScenes and contains labels for one \textit{stuff} class and the same ten \textit{thing} classes as available in Panoptic nuScenes. To match the convention of Panoptic nuScenes, we manually add the labels for the five missing \textit{stuff} classes to a small ($k=30$) subset of images.\footnote{These annotations are available with our code.} Following previous work~\cite{kappeler2024spino, voedisch2025pastel}, we carefully select these images to capture all semantic categories. We then use these human-annotated images to train PASTEL~\cite{voedisch2025pastel}, followed by employing \method to generate panoptic point cloud pseudo-labels for the nuScenes dataset.
Throughout this section, we report results on the \texttt{val} split using the panoptic quality~(PQ) and mean intersection over union~(mIoU) metrics.
\looseness=-1


\begin{table}
\centering
\caption{Panoptic Segmentation Results on nuScenes}
\vspace*{-0.2cm}
\label{table:baseline}
\setlength\tabcolsep{3pt}
\begin{threeparttable}
    \begin{tabular}{l| c| c c}
        \toprule
        \textbf{Method}                       & \textbf{Human annotations} & \textbf{PQ}   & \textbf{mIoU} \\
        \midrule
        \multicolumn{2}{l}{\textit{Fully supervised:}} \\
        [.25ex]
        Panoptic PH-Net~\cite{li2022panoptic} & 28,130 LiDAR scans (\qty{100}{\percent})                & 74.7          & 79.7          \\
        MostLPS~\cite{agarwalla2023lidar}     & ---\textquotedbl---             & 77.1          & 80.3          \\
        P3Former~\cite{xiao2024position}      &---\textquotedbl---             & 75.9          & 76.8          \\
        \grayrule
        \multicolumn{2}{l}{\textit{Semi-supervised:}} \\
        [.25ex]
        ST-SLidR~\cite{mahmod2023self}       & 281 LiDAR scans (\qty{1}{\percent})                  & --             & 40.8          \\
        SEAL~\cite{liu2024seal}               & ---\textquotedbl---          & --             & 45.8          \\
        ScaLR~\cite{puy24scalr}               & ---\textquotedbl---          & --             & 51.0          \\
        \grayrule
        \multicolumn{2}{l}{\textit{Zero-shot:}} \\
        [.25ex]
        SAL\textsuperscript{\textdagger}~\cite{ovsep2025sal} & n/a & 38.4          & 33.9          \\
        \grayrule
        \multicolumn{2}{l}{\textit{Few labels:}} \\
        [.25ex]
        ST-SLidR\textsuperscript{\textdaggerdbl} ~\cite{mahmod2023self} & 30 LiDAR scans (\qty{0.1}{\percent}) & -- & 44.3 \\
        ScaLR\textsuperscript{\textdaggerdbl} ~\cite{puy24scalr} & ---\textquotedbl---  & --   & 58.0 \\
        \method pseudo-labels (\textit{ours}) & 30 images & 48.1 & 56.5 \\
        \method + ScaLR (\textit{ours})       & ---\textquotedbl--- & --   & \textbf{60.1} \\
        \method + P3Former (\textit{ours})    & ---\textquotedbl--- & \textbf{51.4} & 48.2 \\
        \bottomrule
    \end{tabular}
    Segmentation results on the nuScenes~\cite{fong2022panoptic} validation set. The percentage in parentheses is relative to the size of the full training set. The highest metrics among the non-fully supervised approaches are highlighted in bold.
    \textdagger:~SAL uses SAM~\cite{kirillov2023sam} that is trained with human supervision.
    \textdaggerdbl:~Baselines trained by us. The other metrics are from the respective publication.
\end{threeparttable}
\vspace*{-.3cm}
\end{table}


\subsection{Main Results}

In the following, we first present quantitative results comparing our \method approach to various baselines, followed by qualitative results.


{\parskip=2pt
\noindent\textit{Quantitative Results:}
In the first experiment, we compare our \method approach with various existing methods for LiDAR panoptic segmentation, providing evidence for our claims~(1) and~(3). We report results in \cref{table:baseline}, categorized by different levels of supervision.
We list three sets of metrics based on our \method: the quality of the pseudo-labels after 3D refinement, semantic segmentation with ScaLR~\cite{puy24scalr} after fine-tuning with these labels, and panoptic segmentation with P3Former~\cite{xiao2024position} trained with the corresponding panoptic pseudo-labels.
}

First, we present quantitative results of fully supervised LiDAR panoptic segmentation methods, using \qty{100}{\percent} of the annotated point cloud \texttt{train} data. Panoptic PH-Net~\cite{li2022panoptic} uses heatmap clustering, MostLPS~\cite{agarwalla2023lidar} uses a detection-centric approach, and P3Former~\cite{xiao2024position} adopts a transformer architecture. We observe that \method falls short of fully supervised approaches; however, it is important to emphasize that our approach relies on an extremely limited subset of annotated data (30 images vs. 28,130 LiDAR scans), consisting of images rather than point clouds.


\begin{figure*}[t]
    \centering
    \includegraphics[width=.95\linewidth]{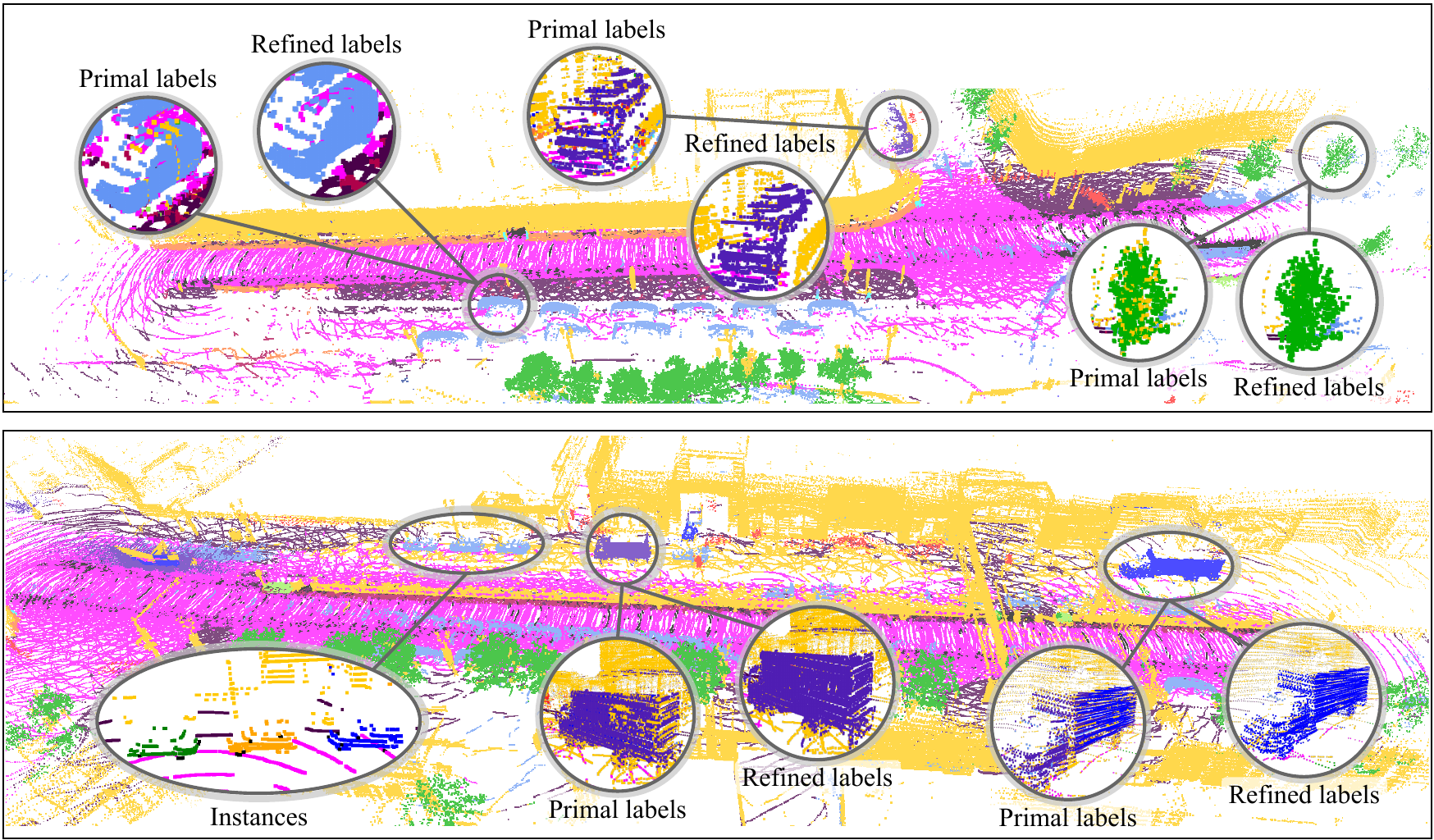}
    \vspace*{-.2cm}
    \caption{Qualitative results on the nuScenes dataset~\cite{fong2022panoptic} generated by our \method approach. We show improvements of the 3D refinement module (``refined labels'') compared to a naive image-to-point-cloud projection (``primal labels''). While we only visualize semantic segmentation in the background map, in the bottom figure we include instance-level pseudo-labels of a single LiDAR scan. Class-wise color legend can be found in~\cref{table:class_wise}.}
    \label{fig:qualitative-results}
\end{figure*}

 
Second, we compare with semi-supervised methods in which \qty{1}{\percent} of the \texttt{train} split is commonly used to assess performance. In particular, we evaluate \mbox{ST-SLidR}~\cite{mahmod2023self} that uses self-supervised learning with superpixels, SEAL~\cite{liu2024seal} that distills multiple foundation models, and ScaLR~\cite{puy24scalr} that leverages pretraining on multiple datasets. In \cref{table:baseline}, we include the results as reported by the respective method. Our approach (``\method + ScaLR'') instead relies on only 30 annotated images and achieves $+19.3$~mIoU over ST-SLidR, $+14.3$~mIoU compared to SEAL, and $+9.1$~mIoU compared to ScaLR. Furthermore, we also compare our method with a recently proposed zero-shot panoptic segmentation model SAL~\cite{ovsep2025sal}, which employs SAM~\cite{kirillov2023sam} for generalized segmentation without fine-tuning. Our method (``\method + P3Former'' and ``\method + ScaLR'') achieves substantial gains of $+13.0$~PQ and $+26.2$~mIoU with only 30 annotated images. This underscores the advantage of \method in scenarios with minimal supervision.

Third, we compare our method with label-efficient methods, which are paramount for large-scale datasets. We also fine-tune ST-SLidR and ScaLR using 30 annotated point clouds for a fair comparison. These 30 samples correspond to approximately \qty{0.1}{\percent} of the full \texttt{train} split and are randomly sampled while ensuring that all classes are represented. We repeat this procedure for five different subsets and average the results from five-fold fine-tuning of \mbox{ST-SLidR} and ScaLR. Our method (``\method + ScaLR'') outperforms \mbox{ST-SLidR} by $+15.8$~mIoU and ScaLR by $+2.1$~mIoU, while requiring only image ground truth annotations. This demonstrates the cost-effectiveness of using image annotations over point cloud labels. 
\looseness=-1

Consequently, the results indicate that by leveraging 2D annotations, \method optimizes the balance between annotation efficiency and segmentation performance, positioning it as a viable alternative for label-efficient LiDAR segmentation and supporting our claim of reducing the reliance on expensive point cloud annotations. Furthermore, the results demonstrate that \method achieves competitive performance relative to existing baselines while requiring less annotation effort.


\begin{table}[t]
\centering
\caption{Ablation Study: 3D Refinement}
\vspace*{-0.2cm}
\label{table:3drefinement}
\setlength\tabcolsep{6pt}
\begin{threeparttable}
    \begin{tabular}{l | cc | cc | cc}
        \toprule
        \textbf{Method} & \multicolumn{2}{c|}{\textbf{Pseudo-labels}} & \multicolumn{2}{c|}{\textbf{ScaLR}}  & \multicolumn{2}{c}{\textbf{P3Former}} \\
        & PQ & mIoU & PQ & mIoU & PQ & mIoU \\
        \midrule
        Primal labels & 37.5 & 48.6 & -- & 51.2 & 27.5 & 28.7 \\
        \hspace{.5pt} + 3D refinement & 48.1 & 56.5 & -- & 60.1 & 51.4 & 48.2 \\
        \grayrule
        & +10.6 & +7.9 & -- & +8.9 & +23.9 & +19.5 \\
        \bottomrule
    \end{tabular}
    We analyze the effect of our proposed 3D refinement module on the vanilla pseudo-labels as well as on using them to train ScaLR~\cite{puy24scalr} for semantic segmentation and P3Former~\cite{xiao2024position} for panoptic segmentation.
\end{threeparttable}
\end{table}


{\parskip=2pt
\noindent\textit{Qualitative Results:}
We visually demonstrate the performance of \method in~\cref{fig:qualitative-results}. We illustrate how our 3D refinement module enhances segmentation quality by improving label consistency, reducing noise, and achieving better instance separation. Compared to primal labels, the refined labels show clearer object boundaries and a more structured representation of the scene. By leveraging geometric properties and temporal accumulation, our approach ensures more accurate and coherent panoptic segmentation, reinforcing the effectiveness of \method in a label-efficient setting.
}


\subsection{Ablation Studies}
\label{subsec:ablation}

In the following ablation studies, we first evaluate the efficacy of our proposed 3D refinement post-processing, supporting claim~(2). Afterward, we analyze the impact of the individual components of \method and provide a study on the parameter sensitivity of the clustering step.


\begingroup
\renewcommand{\arraystretch}{1.1}
\begin{table*}[t]
\centering
\caption{Ablation Study: Class-wise Performance}
\vspace*{-0.2cm}
\label{table:class_wise}
\setlength{\tabcolsep}{0.5em}
\begin{threeparttable}
    \resizebox{0.85\textwidth}{!}{%
    \begin{tabular}{l|c|c|cccccccccccccccc}
    \multicolumn{2}{l|}{\textbf{Method}} & \begin{turn}{90}Mean\end{turn} & 
    \begin{turn}{90}\textcolor{barrier}{\ding{108}} Barrier\end{turn} & 
    \begin{turn}{90}\textcolor{bicycle}{\ding{108}} Bicycle\end{turn} & 
    \begin{turn}{90}\textcolor{bus}{\ding{108}} Bus\end{turn} & 
    \begin{turn}{90}\textcolor{car}{\ding{108}} Car\end{turn} & 
    \begin{turn}{90}\textcolor{construction}{\ding{108}} Constr. veh.\end{turn} & 
    \begin{turn}{90}\textcolor{motorcycle}{\ding{108}} Motorcycle\end{turn} & 
    \begin{turn}{90}\textcolor{pedestrian}{\ding{108}} Pedestrian\end{turn} & 
    \begin{turn}{90}\textcolor{traffic}{\ding{108}} Traffic cone\end{turn} & 
    \begin{turn}{90}\textcolor{trailer}{\ding{108}} Trailer\end{turn} & 
    \begin{turn}{90}\textcolor{truck}{\ding{108}} Truck\end{turn} & 
    \begin{turn}{90}\textcolor{driveable}{\ding{108}} Driv. space\end{turn} & 
    \begin{turn}{90}\textcolor{otherflat}{\ding{108}} Other flat\end{turn} & 
    \begin{turn}{90}\textcolor{sidewalk}{\ding{108}} Sidewalk\end{turn} & 
    \begin{turn}{90}\textcolor{terrain}{\ding{108}} Terrain\end{turn} & 
    \begin{turn}{90}\textcolor{manmade}{\ding{108}} Manmade\end{turn} & 
    \begin{turn}{90}\textcolor{vegetation}{\ding{108}} Vegetation\end{turn} \\
    \toprule
    \multirow{2}{*}{Primal labels} & PQ & 37.5 & 28.5 & 24.8 & 39.4 & 42.0 & 26.3 & 46.9 & 38.6 & 45.0 & 11.4 & 23.7 & 79.5 & 6.6 & 25.6 & 35.1 & 62.9 & 63.9 \\
    & mIoU & 48.6 & 51.3 & 25.4 & 35.4 & 68.7 & 35.7 & 49.3 & 51.8 & 41.0 & 24.9 & 48.7 & 81.2 & 15.5 & 45.7 & 59.7 & 71.4 & 71.1 \\
    \grayrule
    \multirow{2}{*}{3D refinement (\method)}                & PQ & 48.1 & 29.1 & 64.1 & 37.2 & 46.3 & 47.4 & 65.1 & 63.4 & 67.5 & 17.0 & 29.3 & 83.2 & 7.9 & 30.7 & 37.6 & 71.4 & 72.9 \\
    & mIoU & 56.5 & 60.0 & 40.4 & 43.0 & 75.4 & 50.3 & 62.6 & 71.1 & 52.8 & 28.8 & 53.4 & 84.7 & 19.1 & 48.0 & 61.8 & 76.5 & 76.7 \\
    \bottomrule
    \end{tabular}
    }
    \vspace{5pt}
    \\ Class-wise segmentation results on the nuScenes~\cite{fong2022panoptic} validation set for both primal and refined labels.
\end{threeparttable}
\vspace*{-.3cm}
\end{table*}
\endgroup

\begin{table}
\centering
\caption{Ablation Study: Components Analysis}
\vspace*{-0.2cm}
\label{table:component}
\setlength\tabcolsep{12pt}
\begin{threeparttable}
    \begin{tabular}{l | c@{\hspace{2pt}}r c@{\hspace{2pt}}r}
        \toprule
        \textbf{Method} & \multicolumn{2}{c}{\textbf{PQ}} & \multicolumn{2}{c}{\textbf{mIoU}} \\
        \midrule
        Primal labels & 37.5 && 48.6 \\ 
        \hspace{.5pt} + In-cluster majority voting & 47.6 & (+10.1) & 54.5 & (+5.9) \\
        \hspace{.5pt} + Ground separation & 48.0 & (+0.4) & 55.9 & (+1.4) \\
        \hspace{.5pt} + Favoring rare classes & \textbf{48.1} & (+0.1) & \textbf{56.5} & (+0.6) \\
        \bottomrule
    \end{tabular}
    We analyze the impact of the steps of our 3D refinement module. Starting with the primal labels, we initially employ vanilla majority voting within each cluster. Next, we add ground and non-ground partition separations. Finally, we customize the voting mechanism by prioritizing rare classes.
\end{threeparttable}
\vspace*{-.5cm}
\end{table}

{\parskip=3pt
\noindent\textit{3D Refinement:}
We report results on the \texttt{val} split before and after employing our proposed 3D~refinement module in \cref{table:3drefinement}. Projecting raw 2D~predictions onto point clouds results in what we refer to as primal labels (see \cref{ssec:primal-label-generation}). This initial step achieves a performance of \qty{37.5}{\percent}~PQ and \qty{48.6}{\percent}~mIoU. After further processing using our 3D~refinement module (see \cref{ssec:3d_refinement}), the label quality improves substantially by $+10.3$~PQ and $+6.6$~mIoU. We report further class-wise results in \cref{table:class_wise}.
Next, we evaluate the module's impact on the performance of the employed downstream models ScaLR~\cite{puy24scalr} and P3Former~\cite{xiao2024position}, i.e., generating pseudo-labels for training with and without 3D~refinement. While ScaLR achieves a validation performance of \qty{51.2}{\percent}~mIoU when being trained with the primal labels, using the refined labels yields a notable improvement of $+7.9$~mIoU. We make similar observations for P3Former. Training with the primal labels results in \qty{27.5}{\percent}~PQ and \qty{28.7}{\percent}~mIoU, which are substantially increased after 3D~refinement by $+24.1$~PQ and $+15.9$~mIoU. We attribute this pronounced effect to the fact that our design choices of the 3D~refinement module explicitly target the more challenging task of panoptic segmentation.
\looseness=-1
}


{\parskip=3pt
\noindent\textit{Components Analysis:} 
In \cref{table:component}, we present a detailed component-wise analysis of our approach. Initially, reassigning semantic classes within point-based clusters, as opposed to using the primal labels, yields an improvement of $+10.1$~PQ and $+5.9$~mIoU. This result clearly demonstrates the substantial impact of refining semantic assignments through spatial clustering. Subsequently, employing ground segmentation, i.e., segregating ground points from non-ground points prior to clustering, provides an additional increase of $+0.4$~PQ and $+1.4$~mIoU. This finding further underscores the importance of isolating ground elements to enhance overall clustering performance. Finally, employing a majority voting scheme that favors rare classes further enhances the pseudo-label quality by $+0.1$~PQ and $+0.6$~mIoU.
}

\begin{table}
\centering
\caption{Ablation Study: Minimum Cluster Size}
\vspace*{-0.2cm}
\label{table:clustersize}
\setlength\tabcolsep{12pt}
\begin{threeparttable}
    \begin{tabular}{c | c | cc}
        \toprule
        \textbf{Cluster size} & \textbf{Runtime} & \multicolumn{2}{c}{\textbf{Pseudo-labels}} \\
        && PQ & mIoU \\
        \midrule
        3 & \qty{62.2}{\second} & 46.8 & 55.9 \\
        \rowcolor{Gray}
        5 & \qty{20.8}{\second} & \textbf{48.1} & \textbf{56.5} \\
        10 & \qty{7.6}{\second} & 47.8 & 55.2 \\
        20 & \qty{5.3}{\second} & 45.9 & 52.9 \\
        50 & \qty{5.0}{\second} & 40.6 & 48.7 \\
        \bottomrule
    \end{tabular}
    The minimum cluster size used in HDBSCAN~\cite{campello2013density}.
    The parameter used in the other experiments is highlighted in gray. The highest scores per metric are printed in bold.
\end{threeparttable}
\vspace*{-.2cm}
\end{table}


{\parskip=3pt
\noindent\textit{Cluster Size:}
In this ablation study, we investigate the impact of varying the minimum cluster size $s$ used in HDBSCAN~\cite{campello2013density} on the quality of the pseudo-labels. We report the PQ and mIoU metrics for $s \in \{3, 5, 10, 20, 50\}$ in \cref{table:clustersize}. Additionally, we measure the average runtime for processing a single scene. As expected, a larger minimum cluster size results in lower runtime due to the creation of fewer clusters. The chosen parameter $s = 5$ provides a reasonable trade-off, yielding the highest scores for both PQ and mIoU metrics while maintaining efficient runtime.
}

%% file: sections/5_conclusion.tex
\section{Conclusion}

In this work, we addressed the challenge of reducing the annotation effort for LiDAR panoptic segmentation. We proposed the novel method \method, which transfers recent label reduction techniques from the image domain to the point cloud domain. Our \method approach consists of two steps, namely (1) leveraging a highly label-efficient method for visual panoptic segmentation to generate image pseudo-labels that are projected onto 3D LiDAR scans and (2) refining these pseudo-labels based on spatial 3D information. We show that the refined point pseudo-labels can be used to train an off-the-shelf network for LiDAR panoptic segmentation, achieving competitive performance compared to various baselines while shifting the annotation effort to the less complex image space.
We believe that our work represents an important step towards cost-efficient point cloud segmentation, contributing to robotic 3D perception. Future research could explore integrating our approach with unsupervised domain adaptation techniques and further advancing point cloud representation learning, drawing parallels to recent progress in the computer vision domain.
\looseness=-1